\documentclass{article}

\PassOptionsToPackage{numbers, compress}{natbib}

\usepackage[preprint]{neurips_data_2024}

\usepackage[utf8]{inputenc} 
\usepackage[T1]{fontenc}    
\usepackage{hyperref}       
\usepackage{url}            
\usepackage{booktabs}       
\usepackage{amsfonts}       
\usepackage{nicefrac}       
\usepackage{microtype}      
\usepackage{xcolor}         
\usepackage{multirow}
\usepackage{graphicx}
\usepackage{subcaption}
\usepackage{pifont}

\title{\textit{MindBench}: A Comprehensive Benchmark for Mind Map Structure Recognition and Analysis}

\author{
  Lei Chen \\
  Meituan \\
  Beijing, China 100012 \\
  \texttt{leichen1997@outlook.com}
  \And
  Feng Yan \\
  Meituan \\
  Beijing, China 100012 \\
  \texttt{yanfeng05@meituan.com}
  \And
  Yujie Zhong \\
  Meituan \\
  Beijing, China 100012 \\
  \texttt{zhongyujie@meituan.com}
  \And
  Shaoxiang Chen \\
  Meituan \\
  Beijing, China 100012 \\
  \texttt{chenshaoxiang@meituan.com}
  \And
  Zequn Jie$^{\dag}$ \\
  Meituan \\
  Beijing, China 100012 \\
  \texttt{jiezequn@meituan.com}
  \And
  Lin Ma$^{\dag}$ \\
  Meituan \\
  Beijing, China 100012 \\
  \texttt{forest.linma@gmail.com}
}

\begin{document}

\maketitle

\begin{abstract}
  Multimodal Large Language Models (MLLM) have made significant progress in the field of document analysis. Despite this, existing benchmarks typically focus only on extracting text and simple layout information, neglecting the complex interactions between elements in structured documents such as mind maps and flowcharts. To address this issue, we introduce the new benchmark named MindBench, which not only includes meticulously constructed bilingual authentic or synthetic images, detailed annotations, evaluation metrics and baseline models, but also specifically designs five types of structured understanding and parsing tasks. These tasks include full parsing, partial parsing, position-related parsing, structured Visual Question Answering (VQA), and position-related VQA, covering key areas such as text recognition, spatial awareness, relationship discernment, and structured parsing. Extensive experimental results demonstrate the substantial potential and significant room for improvement in current models' ability to handle structured document information. We anticipate that the launch of MindBench will significantly advance research and application development in structured document analysis technology. MindBench is available at: \url{https://miasanlei.github.io/MindBench.github.io/}.
\end{abstract}

{\let\thefootnote\relax\footnotetext{$^{\dag}$ Corresponding authors.}}
\section{Introduction}
The rise of Multimodal Large Language Models (MLLM) ~\cite{yin2023survey} has marked the pivotal turning point in the development of artificial intelligence technology. These models, by integrating multiple modalities such as text, vision, and speech, have demonstrated exceptional capabilities in understanding and generating complex content ~\cite{zhu2023minigpt,liu2023improved, li2023blip, bai2023qwen, ye2023mplug, li2023monkey, chen2023internvl}, particularly in the field of document analysis ~\cite{zhang2023llavar,ye2023mplugdoc,ye2023ureader,hong2023cogagent,liu2024textmonkey}, where they significantly enhance the accuracy of information extraction and content comprehension. However, the benchmarks currently used to evaluate these models often focus primarily on extracting text ~\cite{liu2019curved,ch2017total,yuan2022syntax} and simple layout information ~\cite{park2019cord,jaume2019funsd,huang2019icdar2019,svetlichnaya2020deepform,stanislawek2021kleister}, such as positional relationships in tables ~\cite{chi2019complicated,long2021parsing} and invoices ~\cite{riba2019table}, yet frequently overlook the complex interactions between elements in structured documents. This limitation in evaluation hinders our ability to fully understand and assess models in complex real-world scenarios.

In structured documents, interactions between elements are not only manifested through semantics and positioning but also heavily depend on graphical elements such as arrows and brackets. Mind maps, as a common format, effectively organize and display complex information through their unique structures, making the integration and understanding of information more intuitive and efficient. With the advancements in software like XMind and MindManager, the demand for automated processing of these documents has continually increased. Concurrently, this has introduced new challenges to technology, where the tasks involve not only accurately identifying and parsing textual information but more crucially, recognizing the complex relationships between elements. Therefore, developing a comprehensive and practical benchmark for structured document analysis has become particularly urgent. Such a benchmark would not only thoroughly evaluate the performance of models but also inspire the research community to delve deeper into the complex issues of structured document analysis and seek corresponding solutions.

\begin{figure}[t]
    \centering
    \includegraphics[width=1.\linewidth]{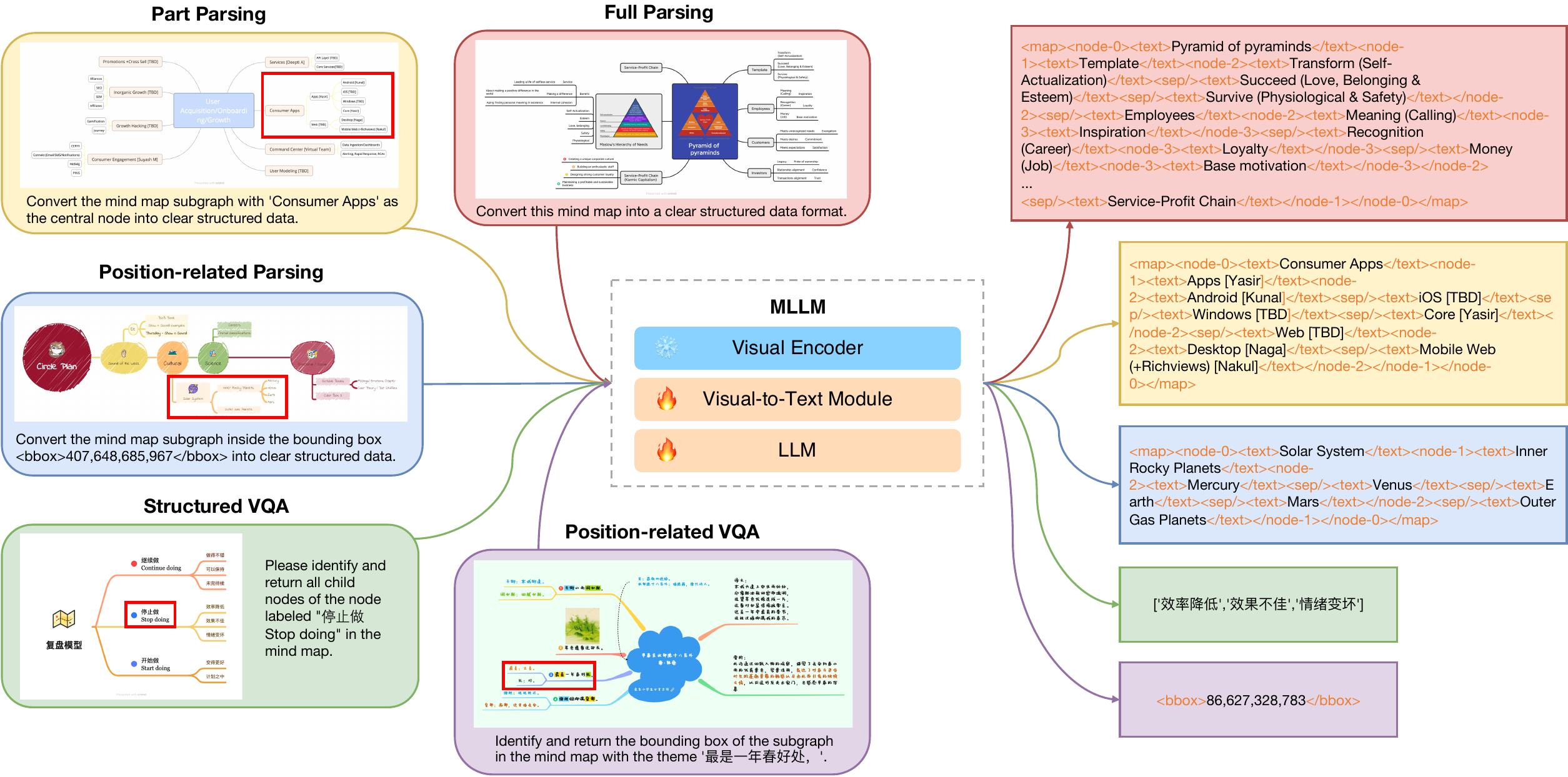}
    \vspace{1mm}
    \caption{The illustration of unified structure learning of the MindBench benchmark.}
    \label{fig:illustration}
\end{figure}

To address the shortcomings of existing benchmarks, this paper introduces a new benchmark called MindBench, specifically designed for the structural analysis and parsing of mind maps. We construct a bilingual dataset of mind maps with high-resolution images, rich document content, and diverse structural variations by parsing the source files of real mind maps and automatically synthesizing simulated mind maps. Based on this dataset, we meticulously design five structured understanding and parsing tasks, as illustrated in Fig.~\ref{fig:illustration}, including full parsing, partial parsing, position-related parsing, structured visual question answering (VQA), and position-related VQA. These tasks comprehensively assess the models' abilities to parse text and image information, recognize relationships between elements, and understand the overall structure. Additionally, we establish specific evaluation metrics, including field-level F1 scores ~\cite{xu2020layoutlm} and Tree Edit Distance (TED)-based accuracy ~\cite{Zhang_Shasha_1989} for parsing tasks, and F1 scores for VQA tasks. Extensive experimental results indicate that there is significant room for improvement in current models, particularly in processing high-resolution complex graphical images and handling lengthy structured document information. This benchmark is expected to significantly advance research and application development in this field.

Our main contributions are as follows:
1. We propose a new benchmark, MindBench, which to our knowledge, is the first benchmark specifically for the analysis of structured documents.
2. This benchmark includes a vast collection of structured document images and corresponding annotation data, along with accompanying evaluation metrics, providing a standardized tool for research in this area.
3. Utilizing this dataset, we train and test several leading models related to this field. The results show that although there has been progress in handling high-resolution complex graphical images and lengthy structured document information, there is still significant potential for improvement.

\section{Related Work}

{\flushleft \bf Visual Document Understanding (VDU)} aims to comprehend text-rich images covering a wide range of types including documents~\cite{mathew2021docvqa,svetlichnaya2020deepform,stanislawek2021kleister,mathew2022infographicvqa}, tables~\cite{pasupat2015compositional,chen2019tabfact,zhong2020image}, charts~\cite{kafle2018dvqa,methani2020plotqa,masry2022chartqa,kantharaj2022chart,tang2023vistext,hu2023mplug}, natural images~\cite{sidorov2020textcaps,singh2019towards,hu2021question}, and screenshots~\cite{tanaka2021visualmrc,chen2021websrc}. The tasks of VDU are diverse, encompassing visual question answering~\cite{mathew2021docvqa,mathew2022infographicvqa,pasupat2015compositional,masry2022chartqa,tanaka2021visualmrc,singh2019towards}, image captioning~\cite{sidorov2020textcaps,kantharaj2022chart,tang2023vistext}, information extraction~\cite{svetlichnaya2020deepform,stanislawek2021kleister} and natural language inference~\cite{chen2019tabfact}. However, the tasks of extraction and understanding for complex structured documents, such as mind maps, have not been taken into consideration. 
Models designed for VDU can be broadly categorized into two types: OCR-model-driven methods and OCR-free methods. OCR-model-driven methods~\cite{xu2020layoutlm,xu2020layoutlmv2,huang2022layoutlmv3,tang2023unifying,appalaraju2024docformerv2,wang2023docllm} use the models to integrate visual data with detected text and layout information from off-the-shelf OCR models. OCR-free methods~\cite{kim2022ocr,davis2022end,lee2023pix2struct,mao2024visually,wan2024omniparser} learn text-layout recognition with a high-resolution image encoder in an end-to-end manner. Both of these VDU methods require fine-tuning for specific tasks.

{\flushleft \bf Multimodal Large Language Models (MLLM)} have recently been developed for general visual language understanding~\cite{zhu2023minigpt,liu2023improved, li2023blip, bai2023qwen, ye2023mplug, li2023monkey, chen2023internvl}, leveraging the powerful language comprehension and general capabilities of Large Language Models (LLM)~\cite{brown2020language,touvron2023llama,vicuna,zhao2023survey}. These approaches utilize a common architectural paradigm that connects a visual encoder, e.g., ViT~\cite{dosovitskiy2020image,radford2021learning} to a Large Language Model through a visual-to-text module, e.g., linear layers~\cite{liu2024visual,liu2023improved} or a Q-Former~\cite{bai2023qwen}/Resampler~\cite{alayrac2022flamingo}/Abstractor~\cite{ye2023mplug,ye2023mplug2} with learnable queries. To facilitate the comprehension of text-rich images by MLLMs, several research efforts~\cite{zhang2023llavar,ye2023mplugdoc,feng2023unidoc} have explicitly conducted tuning instructions on visual text understanding datasets. To handle high-resolution document images, some methods~\cite{ye2023ureader,hu2024mplug,dong2024internlm4k,chen2024far} employ shape-adaptive cropping modules to segment images into resolutions suitable for ViT models. Additionally, to enhance the understanding of document text and structured information, various tasks such as text reading~\cite{bai2023qwen,ye2023ureader,dong2024internlm4k,chen2024far}, text grounding~\cite{bai2023qwen,hong2023cogagent,feng2023docpedia,hu2024mplug,liu2024textmonkey}, and table parsing~\cite{hu2024mplug,liu2024textmonkey} have been designed. However, these tasks primarily focus on learning text recognition and simple layout information, overlooking the complex interactions among elements in structured documents. In this paper, we introduce a comprehensive benchmark called MindBench for structured document parsing and understanding. This benchmark allows for the evaluation of various capabilities of existing models, encompassing document text recognition, layout information perception, and complex interaction understanding.

\section{The MindBench Dataset}
\subsection{Data Generation}

{\flushleft \bf Data preparation.} Given the limited availability of labeled mind map data online, we synthesize additional mind map data using a multi-step process. Firstly, we randomly sample textual content of the nodes. Then, we generate mind maps in various shapes by randomly sampling the number of nodes, node children, and depths. These structured mind maps are then rendered into images using the Graphviz tool. To ensure diversity, we incorporate various layout engines and a wide range of properties for nodes and edges. Furthermore, we randomly place 0 to multiple background images and apply Gaussian noise to bring background diversity. The synthetic examples are shown in Fig.~\ref{fig:sft_infer}. While synthesizing data, we also recognize the importance of validating the models on real-world data. Hence, we make efforts to download a limited number of mind map source files from open-source mind map websites, including XMind\footnote{\url{https://xmind.app/share/},\url{https://xmind.cn/mindmaps-gallery/}}, Biggerplate\footnote{\url{https://www.biggerplate.com/mindmap-library}}, and Zhixi\footnote{\url{https://www.zhixi.com/space\#src=btn}}.

\begin{figure}[t]
    \centering
    \includegraphics[width=1.\linewidth]{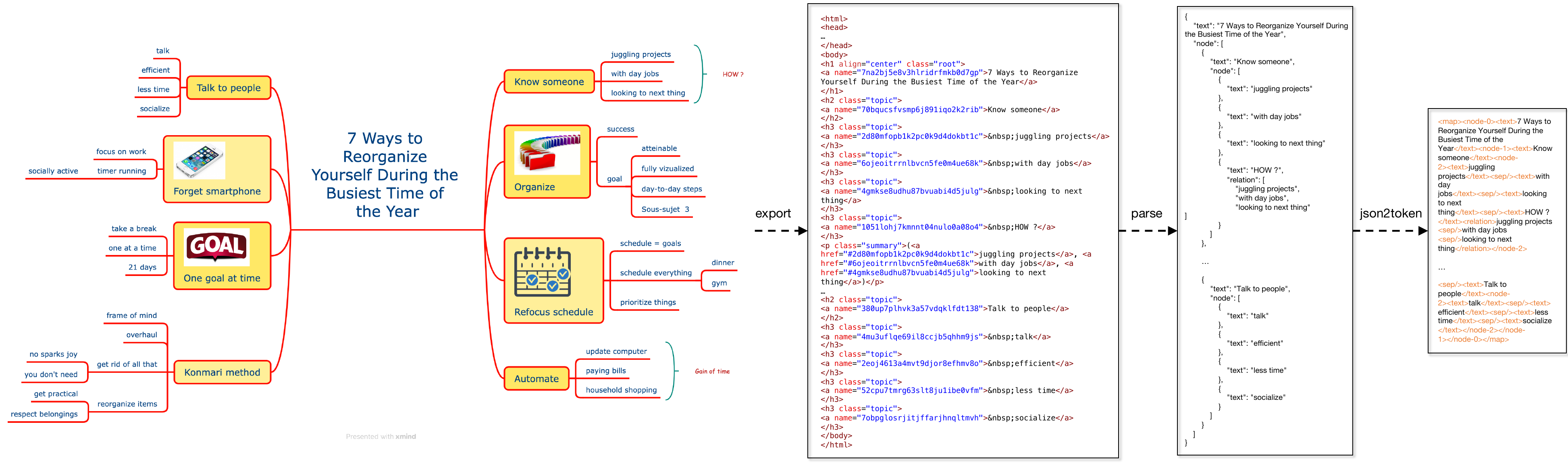}
    \vspace{1mm}
    \caption{{The illustration of data parsing.}}
    \label{fig:parse_procedure}
\vspace{-3mm}
\end{figure}

{\flushleft \bf Data parsing.} In order to obtain unified structured annotations for training and evaluation, we parse the raw files of two types of data, preserving the textual and structural information while removing redundant information. Fig.~\ref{fig:parse_procedure} illustrates the parsing process for the crawled data. First, we use the XMind software to automate the export of PNG images and HTML tag files of the source files. The HTML file contains structured information about the mind map. Then, we employ BeautifulSoup to parse the HTML, maintaining the tree structure and relationships among nodes, and convert the mind map into a nested JSON format. In the JSON structure, the node's children were represented as a list, allowing for nested nodes. For training, we convert the JSON data into a token sequence, ensuring reversibility by adding hierarchical sequence numbers to nested nodes. To avoid confusion with existing special tokens in MLLMs, we prefix all attribute names with `s\_'. For the synthetic data, we directly convert the generated tree structure of the mind map into a token sequence, ensuring consistency with the labeling format of the crawled data.

\subsection{Task Definition}
Fig.~\ref{fig:illustration} illustrates five OCR-free tasks we designed, focusing on mind map structure parsing and understanding, which are elaborated in the following:

{\flushleft \bf Full parsing.} As indicated by the red rectangle in Fig.~\ref{fig:illustration}, the task requires the model to return the full parsing results of the input mind map image, specifically the final token sequence discussed in the previous subsection. Mind map images, as depicted in Fig.~\ref{fig:reso}, often have significantly higher resolutions than typical document images, with some exceeding 10,000 pixels. This demands models capable of processing high-resolution images. However, most existing MLLMs handle only up to 1000 pixels, and even advanced models~\cite{dong2024internlm4k} supporting up to 4k pixels struggle to clearly display text in many nodes. Furthermore, higher resolution mind maps contain more information, resulting in longer structured data, which presents a significant challenge for existing models. We utilize all crawled data and the majority of the synthetic data to perform this task.

{\flushleft \bf Part parsing.} This task involves returning a subgraph centered around a specific node, resulting in shorter token output. This can alleviate pressure on models that struggle with insufficient processing length. However, it also poses new challenges, requiring the model to accurately identify the central theme node from the question and return its corresponding subgraph based on a thorough understanding of the mind map structure. Additionally, this task addresses the tendency of models to parse from the beginning, similar to the rationale behind continue reading task. However, this task does not provide preceding texts but prompts only with the theme name, posing a greater challenge.

{\flushleft \bf Position-related parsing.} Similar to part parsing, this task also returns a subgraph of the mind map. The difference is that this task emphasizes spatial positioning, requiring the model to integrate capabilities in text recognition, spatial awareness, and relational parsing. Since the crawled data's exported HTML lacks coordinate information, this task is conducted on synthetic data, where we can extract the bounding boxes of each node from Graphviz source files. As in previous works~\cite{bai2023qwen,hu2024mplug,liu2024textmonkey}, we describe the bounding box as ``<bbox>x1,y1,x2,y2</bbox>'', normalizing the coordinates to integers between 0 and 999.

{\flushleft \bf Structured VQA.} Besides the parsing tasks, we design multiple VQA sets to enable explicit learning of the components of mind maps and their interrelationships. For instance, we craft prompts such as, ``Describe the central theme of the mind map.'' Typically, the central theme of a conventional mind map is easily identifiable, often located at the center or along the middle of an edge. However, in some layouts, such as the image in Fig.~\ref{fig:pred3}, identifying the central theme is challenging. An initial misprediction of the central node can lead to subsequent structural confusion and parsing failures. Thus, explicitly retrieving the central theme is crucial. We also design VQA tasks related to node kinship and hierarchical relationships, with specific prompts provided in Appendix~\S~B.

{\flushleft \bf Position-related VQA.} We design two types of position-related VQA tasks: recognition and grounding. In recognition tasks, the model receives node coordinates and returns answers about structural information. For example, the instruction ``How many nodes are contained within the bounding box <bbox>[content]</bbox>?'' requires the model for both localization and counting capabilities. In grounding tasks, the model receives node descriptions and returns the bounding box coordinates of the corresponding structure. For example, ``Return the bounding box of the subgraph with the theme '[content]'.'' The model needs to identify the central theme mentioned in the instruction, understand the positional relationships with its descendant nodes, and return the coordinates of the entire subgraph. The coordinates of the subgraph are represented by the minimum and maximum coordinates of all nodes within it. More position-related VQA prompts can be found in Appendix~\S~B.

Overall, the proposed five tasks are designed to enhance model comprehensive capabilities in text detection, relationship recognition, spatial awareness, and structure parsing.

\subsection{Statistic}

\begin{table}[h]
  \caption{Statistics of our MindBench datasets. `BXMind' and `BMManager' represent data downloaded from the Biggerplate website in XMind and MindManager formats, respectively. `EN' and `CN' indicate the language of the data as English and Chinese, respectively. $^{*}$ denotes the number of samples in the test set with fewer than 60 nodes. We use these simpler test sets for validation unless otherwise specified.}
  \label{statistics}
  \centering
  \hspace*{-1cm}
  \begin{subtable}{.58\textwidth}
    \centering
    \caption{Statistics of crawled dataset.}
    \label{crawl_stat}
    \begin{tabular}{lllll}
      \toprule
      \textbf{source}   & \textbf{lang} & \textbf{train}  & \textbf{test} & \textbf{test$^{*}$} \\
      \midrule
      XMind & EN  & 2747  &  145  & 69  \\
      XMind & CN  & 1518  &  169  & 72  \\
      BXMind & EN & 2007  & 224  &  143  \\
      BMManager & EN & 3707  & 412  &  243  \\
      Zhixi & CN  & 11548 &  608  & 342 \\ 
      \midrule
      total & EN+CN & 21527 & 1558 & 869 \\
      \bottomrule
    \end{tabular}
  \end{subtable}
  \begin{subtable}{.42\textwidth}
    \centering
    \caption{Statistics of synthetic dataset.}
    \label{synth_stat}
    \begin{tabular}{llll}
      \toprule
      \textbf{task}   & \textbf{type} & \textbf{train} & \textbf{test} \\
      \midrule
      \multirow{3}{*}{Parse} &  Full  & 200,000 & 100\\
       &  Part  & 40,000 & 100 \\
       &  Position-related  & 40,000 & 100 \\
      \multirow{2}{*}{VQA}   &  Structured & 60,000 & 100 \\
       &  Position-related  & 60,000 & 100 \\
      \midrule
      total & mixed & 400,000 & 500 \\
      \bottomrule
    \end{tabular}
  \end{subtable}
\end{table}

{\flushleft \bf Dataset Splits.} Table~\ref{crawl_stat} displays the data downloaded from multiple websites, segmented into training and testing sets. To accurately assess the model's ability to handle mind maps of varying complexities, we select a subset of simpler data (test$^*$) from the test set based on the crucial metric of node number, which serves as our default validation set. Our research indicates that using large mind maps with a higher number of nodes during the training phase greatly benefits structural parsing learning; therefore, our training set encompasses data of various complexities. Table~\ref{synth_stat} lists the volume of synthetic data used for each task, with the key full parsing task utilizing a larger number of samples, and all synthetic data evenly distributed between English and Chinese. It should be noted that due to the non-uniqueness of node content and the absence of coordinate information in the crawled data, we primarily use this data for full parsing tasks to ensure high data quality. In the future, part parsing and VQA tasks could also consider utilizing this data for further research.

\begin{figure}[t]
    \centering
    \begin{subfigure}[b]{0.32\textwidth}
        \includegraphics[width=\textwidth]{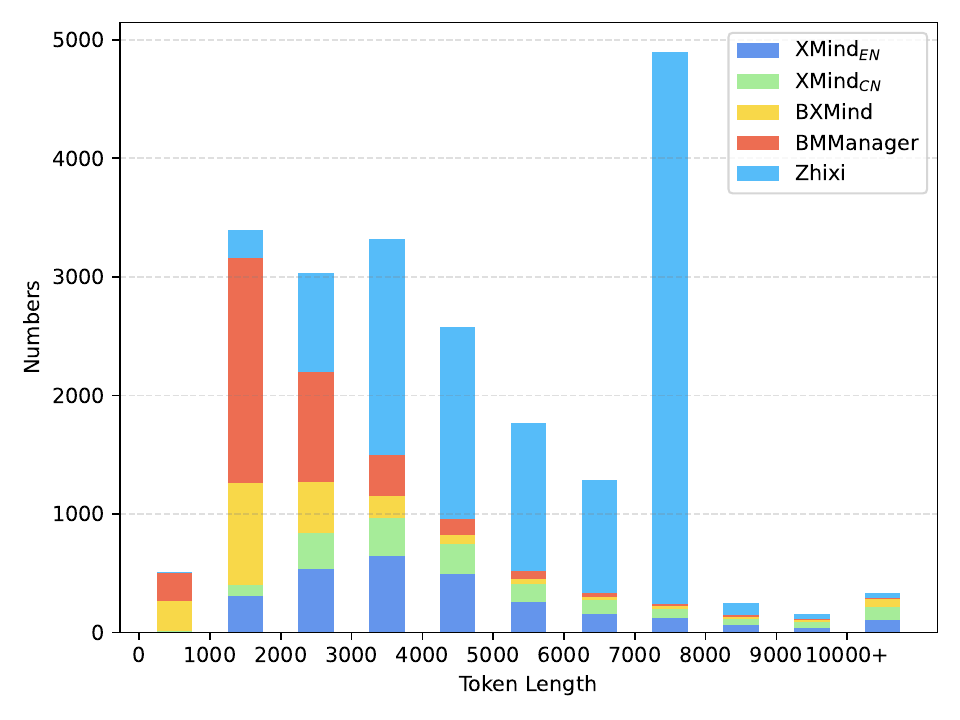}
        \caption{Resolution distribution of crawled dataset.}
        \label{fig:reso}
    \end{subfigure}
    \hfill
    \begin{subfigure}[b]{0.32\textwidth}
        \includegraphics[width=\textwidth]{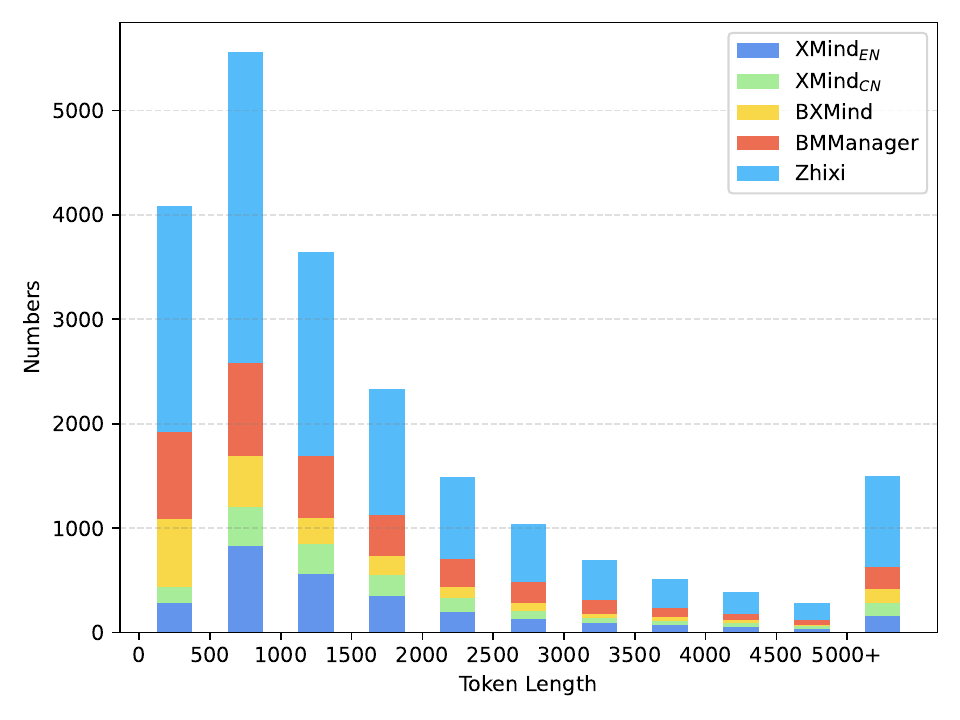}
        \caption{Token length distribution of crawled dataset.}
        \label{fig:token_len1}
    \end{subfigure}
    \hfill
    \begin{subfigure}[b]{0.32\textwidth}
        \includegraphics[width=\textwidth]{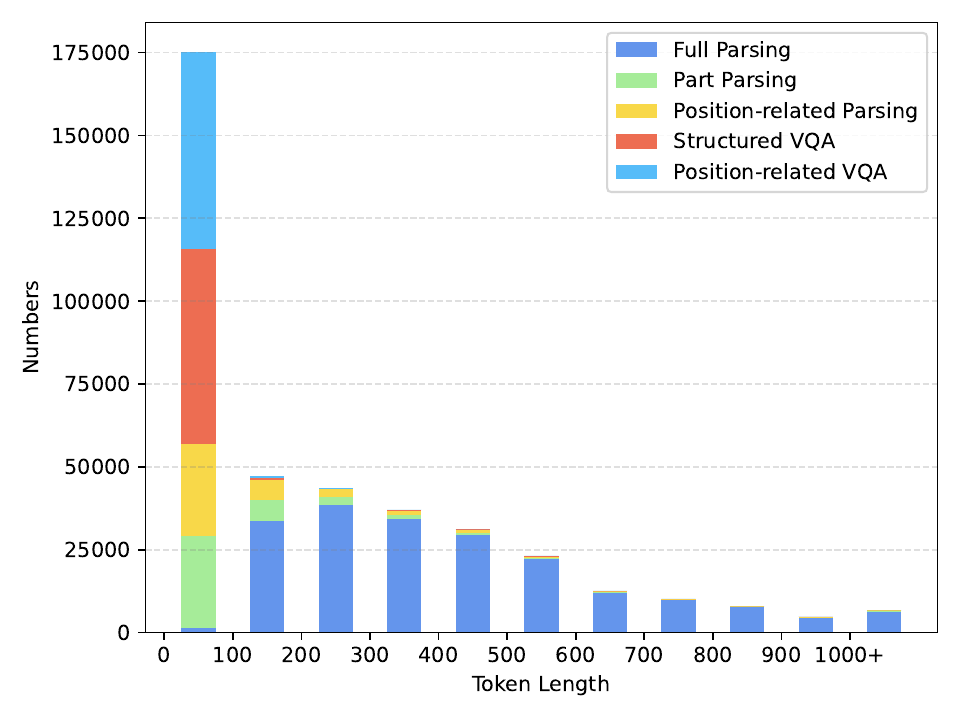}
        \caption{Token length distribution of synthetic dataset.}
        \label{fig:token_len2}
    \end{subfigure}
    \caption{Resolution and token length distributions.}
\end{figure}

{\flushleft \bf Resolution.} The sizes of images are crucial for model processing capabilities, hence we conduct a detailed analysis of the resolution distribution of the crawled data. As depicted in Fig.~\ref{fig:reso}, we present the length of the longest side of images from various sources alongside their corresponding numbers. Among these, BXMind and BMManager feature relatively low resolutions, typically ranging from 1000 to 3000 pixels, while the resolution distribution of XMind exhibits a normal distribution pattern. Notably, Zhixi has higher resolutions, usually between 7000 to 8000 pixels, posing significant challenges to existing MLLMs: when these high-resolution images are scaled down to the input resolutions of the models, the texts often become illegible. As for the synthetic data, its resolution is influenced by the layout engine and the number of nodes. During synthesis, we uniformly sample these two parameters to ensure a consistent resolution distribution across all tasks.

{\flushleft \bf Token length.} Token length is another crucial metric determining the processing capabilities of models. As illustrated in Fig.~\ref{fig:token_len1} and Fig.~\ref{fig:token_len2}, we conduct a detailed analysis of the token length distribution in both crawled and synthetic data. In the crawled data, the token lengths exhibit a long-tail distribution, particularly in samples from Zhixi, where many samples exceed 5000 tokens. This poses a challenge to existing MLLMs, as these models typically have a maximum processing length limited to 4096 tokens, including visual tokens.
In the synthetic data, the token count for VQA responses usually falls below 100 tokens. Compared to full parsing, the token lengths for part parsing and position-related parsing are shorter. Additionally, the token length distribution in synthetic data is more uniform, with fewer extremes compared to the crawled data.

\begin{table}[h]
  \caption{{Structure statistic of different datasets.} Nodes and depth are the average number of samples.}
  \label{attr}
  \centering
  \begin{tabular}{ccccccc}
    \toprule
    & $\textbf{XMind}_\textit{EN}$ & $\textbf{XMind}_\textit{CN}$ & $\textbf{BXMind}$ & $\textbf{BMManager}$ & $\textbf{Zhixi}$ & \textbf{Synthetic} \\
    \midrule
    nodes & 100 & 112 & 94  &  76  &  91 &  16  \\
    depth & 5.6 & 5.8 & 4.7 &  4.5 & 5.6 & 4.8 \\
    \bottomrule
  \end{tabular}
\end{table}

{\flushleft \bf Structure.} To fully understand the structural distribution, Table~\ref{attr} provides detailed information on the number of nodes and depth across different datasets, with XMind and Zhixi exhibiting higher structural complexity, aligning with their resolution distributions.

\section{Experiments}
\subsection{Experimental Setup}
\label{sec:setup}

{\flushleft \bf Model.} We evaluate several visual document understanding models~\cite{openai2023gpt,kim2022ocr,ye2023ureader,liu2024textmonkey,dong2024internlm4k} on the proposed benchmark. The criteria to select a baseline model are as follows: models are pre-trained on an extensive corpus of OCR and document data, they can possess a sufficiently high input resolution, and is capable of handling documents of substantial length. For implementation details of each model, please consult the respective original publications. In this paper, all models use unified structure learning and perform different tasks depending on the prompt. Due to the limited quantity of the crawled data, it is up-sampled 10 times during training to balance the quantity between the two data types. Table~\ref{setting} provides the comparison of model settings. We employ GPT4V~\cite{openai2023gpt} for two-shot inference to examine whether the existing commercial models have the capability of structural graphical parsing. We then utilize one domain-specific model Donut~\cite{kim2022ocr} and three large document models~\cite{ye2023ureader,liu2024textmonkey,dong2024internlm4k} for SFT on our dataset. The training details, largely in line with the original paper, can be found in Appendix~\S~A.

{\flushleft \bf Metric.} For parsing task, following Donut, we evaluate the models using two metrics: field-level F1 score and Tree Edit Distance (TED) based accuracy. We first convert the predicted token sequence to JSON format to recover the tree structure of the graph. The F1 metric flattens the nested JSON into a non-nested format, and then calculates F1 score at each field. F1 can efficiently evaluate the extracted field information, but it cannot exactly measure the structure of the tree.
The TED-based metric is appropriate for evaluating tree-structured documents. Specifically, it uses the Zhang-Shasha (ZSS) algorithm~\cite{Zhang_Shasha_1989} to calculate the nTED between the prediction tree and the answer tree, where $n$ represents the size of the answer tree. The accuracy based on nTED is then computed using the formula $max(1 - nTED, 0)$. For VQA task, we simply evaluate the models with F1 score.

\begin{table}
  \caption{Different settings of OCR-free visual document understanding models.}
  \label{setting}
  \centering
  \begin{tabular}{ccccccc}
    \toprule
    \textbf{Model}   & \textbf{SFT} & \textbf{Size} & \textbf{Trainable} & \textbf{Resolution} & \textbf{Length} \\
    \midrule
    GPT4V~\cite{openai2023gpt}  & \ding{55} & - & - & - & - \\
    Donut~\cite{kim2022ocr}  & \checkmark & 201M & 201M & 2560x1920 & 1536 \\
    UReader~\cite{ye2023ureader}  & \checkmark & 7.1B & 86M & 224x224(x20 crops) & 2048 \\
    TextMonkey~\cite{liu2024textmonkey}  & \checkmark & 9.7B & 7.9B & 896x896  &  2048 \\
    IXC2-4KHD~\cite{dong2024internlm4k}  & \checkmark & 8.6B & 8.6B & 336x336(x25 crops)  &  4096 \\
    \bottomrule
  \end{tabular}
\end{table}

\subsection{Comparison with SOTA MLLMs}

\begin{table}[h]
  \caption{The performance comparison with OCR-free visual document understanding models on crawled dataset. Values in the table represent TED-based accuracy and field-level F1 score, respectively. `N/A' denotes that UReader lacks Chinese recognition capabilities. $^{\dag}$ indicates evaluation on challenging samples in test sets.}
  \label{sota}
  \centering
  \begin{tabular}{cccccc}
    \toprule
    \textbf{Model}  & $\textbf{XMind}_\textit{EN}$ & $\textbf{XMind}_\textit{CN}$ & $\textbf{BXMind}$ & $\textbf{BMManager}$ & $\textbf{Zhixi}$ \\
    \midrule
    GPT4V~\cite{openai2023gpt}  & 38.6 / 43.6 & -  &  38.2 / 37.8  &  29.3 / 30.8 &  -  \\
    Donut~\cite{kim2022ocr}  & 71.3 / 57.9 & 66.5 / 47.3  &  81.7 / \textbf{72.8}  &  77.2 / 62.9 & 79.7 / 50.0 \\
    UReader~\cite{ye2023ureader}  & 33.3 / 23.8 & N/A  &  52.6 / 35.3  & 39.9 / 30.2 & N/A \\
    TextMonkey~\cite{liu2024textmonkey}  & 54.6 / 43.5 &  50.5 / 39.3  &  73.1 / 57.4  &  62.2 / 49.3 & 68.6 / 51.7 \\
    IXC2-4KHD~\cite{dong2024internlm4k}  & \textbf{73.5 / 66.4} & \textbf{75.1 / 61.4}  &  \textbf{84.5} / 72.5 & \textbf{77.3 / 65.4}  & \textbf{82.5 / 66.8} \\
    \midrule
    IXC2-4KHD$^{\dag}$~\cite{dong2024internlm4k} & 45.8 / 27.5 &  37.3 / 21.8  &  54.2 / 33.8 & 48.3 / 32.8 & 36.5 / 17.6 \\
    \bottomrule
  \end{tabular}
\end{table}

We conduct the performance comparison of existing visual document understanding models on the MindBench benchmark, as detailed in Table~\ref{sota}. GPT4V exhibits mediocre performance, indicating challenges for commercial models in parsing complex structured documents such as mind maps. Donut ranks second in parsing performance, significantly outperforming UReader and TextMonkey, and closely approaching the performance of IXC2-4KHD. This underscores the advantages of domain-specific models for parsing tasks. Although MLLMs are versatile, their capability in structured document understanding is not yet exceptional. IXC2-4KHD delivers the best performance, likely due to extensive OCR data pre-training, higher resolution input, and the capability to handle longer token lengths. 
Additionally, we conduct evaluations on challenging test samples. There is a notable accuracy discrepancy between complex samples with over 60 nodes and simpler ones. This highlights that the capabilities of current MLLMs are still limited when it comes to analyzing complex mind maps, particularly in processing high-resolution complex graphical images and ultra-long structured document information. There is an urgent need for further improvement of MLLM technology.

\begin{table}[h]
  \caption{The performance comparison with OCR-free visual document understanding models on English synthetic dataset. For parsing tasks, values in the table represent TED-based accuracy and field-level F1 score. For VQA tasks, values represent F1 score.}
  \label{synth_sota}
  \centering
  \begin{tabular}{c|ccc|cc}
    \toprule
    \multirow{2}{*}{\textbf{Model}} & \multicolumn{3}{c|}{\textbf{Parse}} &  \multicolumn{2}{c}{\textbf{VQA}} \\
    & {Full} & {Part} & {Position-related} & {Structured} & {Position-related} \\
    \midrule
    UReader~\cite{ye2023ureader}  & 29.4 / 5.0 & 65.2 / 35.1 & 22.7 / 7.8 & 66.7 & 38.7 \\
    IXC2-4KHD~\cite{dong2024internlm4k}  & \textbf{81.8 / 59.8} & \textbf{95.0 / 84.5} & \textbf{90.3 / 65.3} & \textbf{92.5} & \textbf{66.7} \\
    \bottomrule
  \end{tabular}
\end{table}

In Table~\ref{synth_sota}, we compare the performance of UReader and IXC2-4KHD across five subtasks involving synthetic data. IXC2-4KHD consistently outperforms UReader in all tasks.
Full parsing has notably lower accuracy than part or position-related parsing, indicating its greater complexity. Additionally, position-related tasks show consistently lower accuracy than other tasks within the same category, highlighting the challenges of integrating structured understanding with spatial perception.

\subsection{Ablation Study}

\begin{table}[h]
  \caption{The impact of unified structure learning on parsing tasks. `SynFP' represents the synthetic data of full parsing task. `SynOther' represents other synthetic data except SynFP.}
  \label{crawl_synth}
  \centering
  \begin{tabular}{c|ccc|cc}
    \toprule
    \multirow{2}{*}{\textbf{Model}}  & \multicolumn{3}{c|}{\textbf{Data}} & \multirow{2}{*}{$\textbf{XMind}_\textit{EN}$} & \multirow{2}{*}{$\textbf{SynFP}_\textit{EN}$} \\
    & Crawled & SynFP & SynOther &  &  \\
    \midrule
    \multirow{3}{*}{UReader~\cite{ye2023ureader}}  & \checkmark &  &  & 28.8 / 22.0 & 6.8 / 0.6 \\
    & \checkmark & \checkmark &  & 31.5 / 22.5 & \textbf{25.8 / 4.7} \\
    & \checkmark & \checkmark & \checkmark & \textbf{32.3 / 23.5} & 21.8 / 3.0 \\
    \bottomrule
  \end{tabular}
\end{table}

{\flushleft \bf Unified structure learning.} We conduct ablation experiments to analyze the impact of unified structure learning, as presented in Table~\ref{crawl_synth}. To expedite the experiments, we use half of the data for this ablation study. Initially, we fine-tune the UReader model on 50\% of the crawled data and evaluate its performance on the XMind test set as well as the synthetic test set. Due to the disparity in graph style, the model struggles on the synthetic test set.
Subsequently, we introduce the full parsing task with synthetic data during training, resulting in improvements on both the XMind and synthetic test sets. This indicates that incorporating synthetic datasets can significantly aid in parsing real mind maps, even in the presence of substantial style differences. Lastly, we integrate all tasks for unified structure learning. We train the model using 50\% of the full parsing task data and 50\% of other task data, maintaining the same total quantity of synthetic data as in the previous experiment. It can be observed that the model continues to show improvements on the XMind test set, highlighting the effectiveness of explicitly learning inter-node relationships and spatial information for comprehensive structure parsing. However, the model's performance slightly decreases on the synthetic test set, which may be attributed to the reduced quantity of synthetic data in the full parsing task.

\begin{figure}[t]
    \centering
    \includegraphics[width=.9\linewidth]{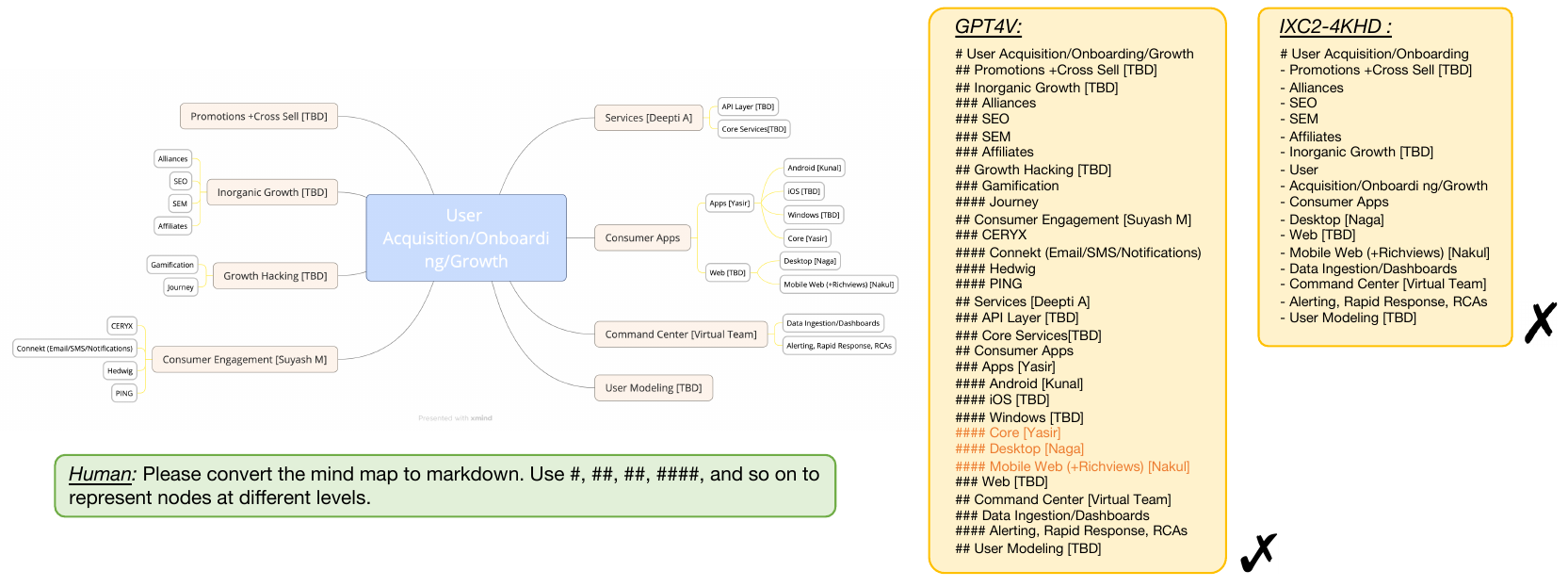}
    \caption{{A qualitative result of existing MLLMs in a zero-shot setting.}}
    \label{fig:zero-shot}
\end{figure}

\begin{figure}[t]
    \centering
    \begin{subfigure}[b]{0.48\textwidth}
        \includegraphics[page=1,width=\textwidth]{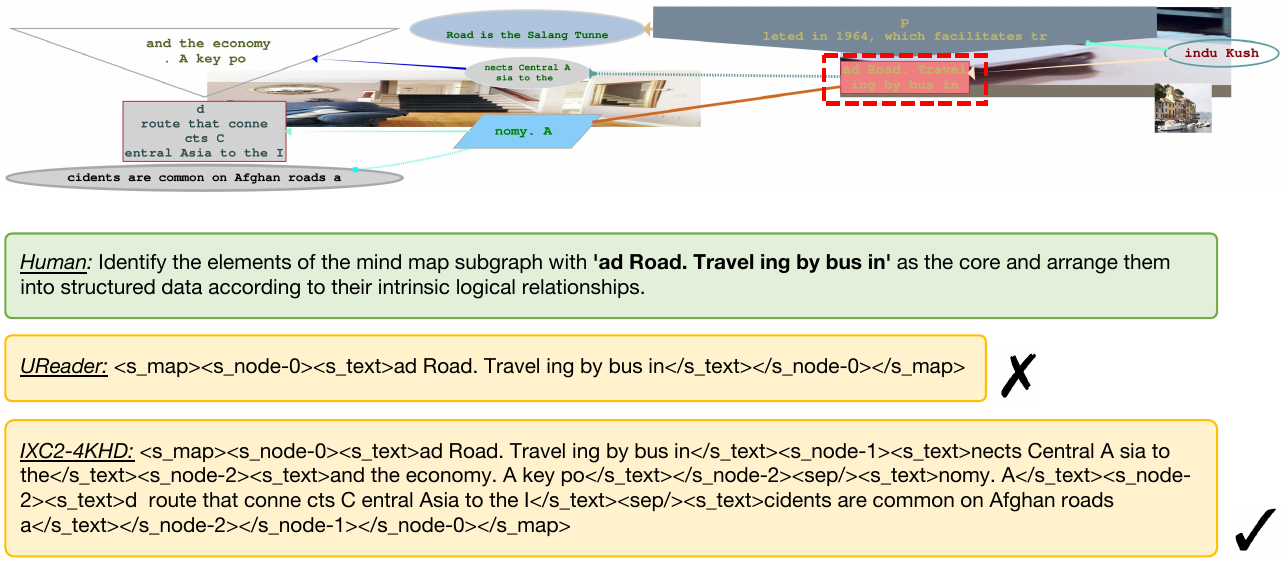}
        \caption{Part parsing}
        \label{fig:pred1}
    \end{subfigure}
    \hfill
    \begin{subfigure}[b]{0.48\textwidth}
        \includegraphics[page=1,width=\textwidth]{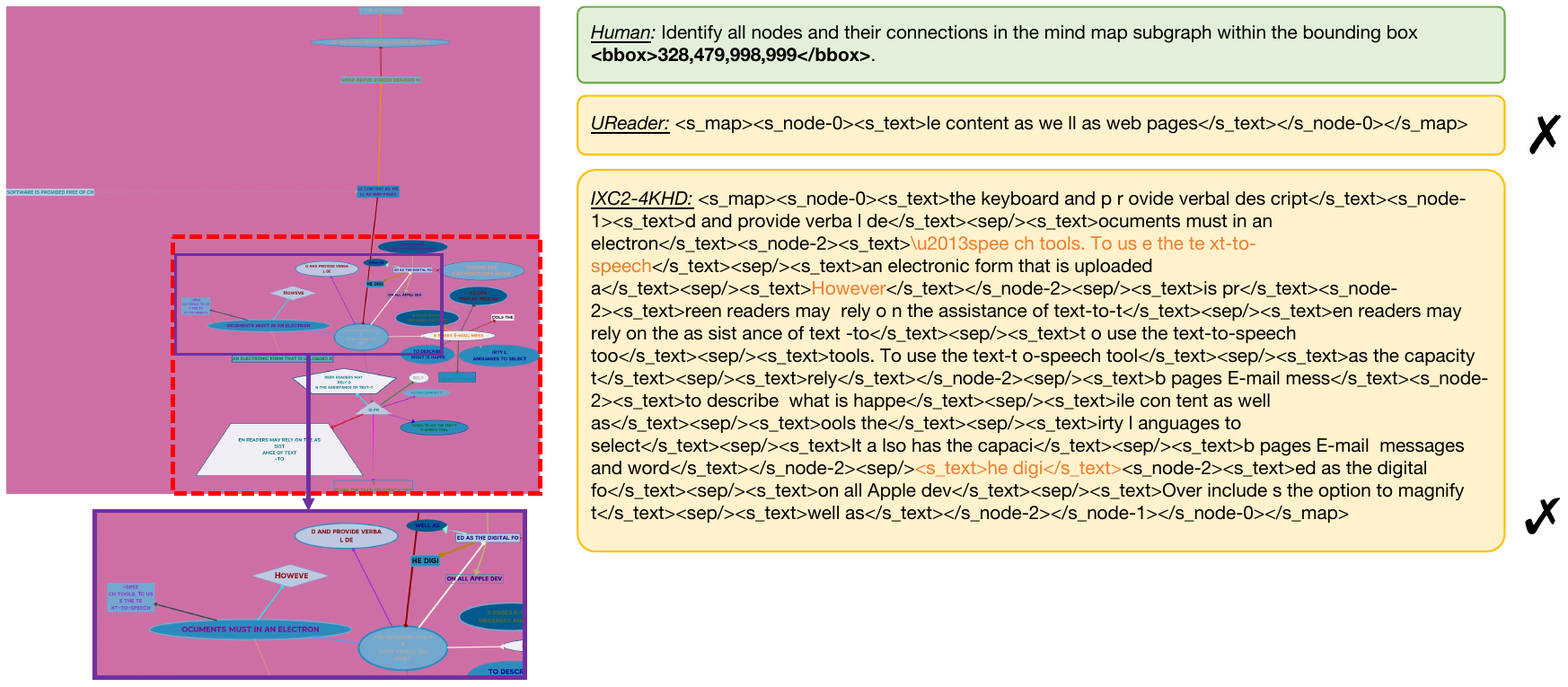}
        \caption{Position-related parsing}
        \label{fig:pred2}
    \end{subfigure}
    \hfill
    \begin{subfigure}[b]{0.48\textwidth}
        \includegraphics[page=1,width=\textwidth]{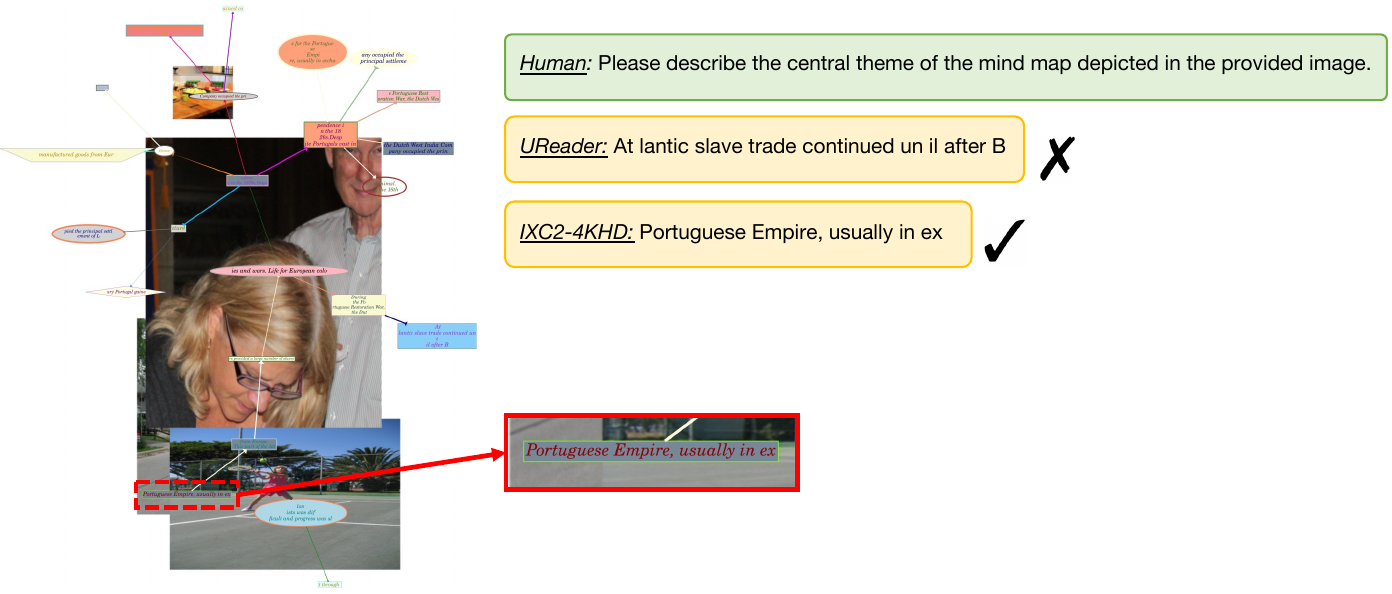}
        \caption{Structured VQA}
        \label{fig:pred3}
    \end{subfigure}
    \hfill
    \begin{subfigure}[b]{0.48\textwidth}
        \includegraphics[page=1,width=\textwidth]{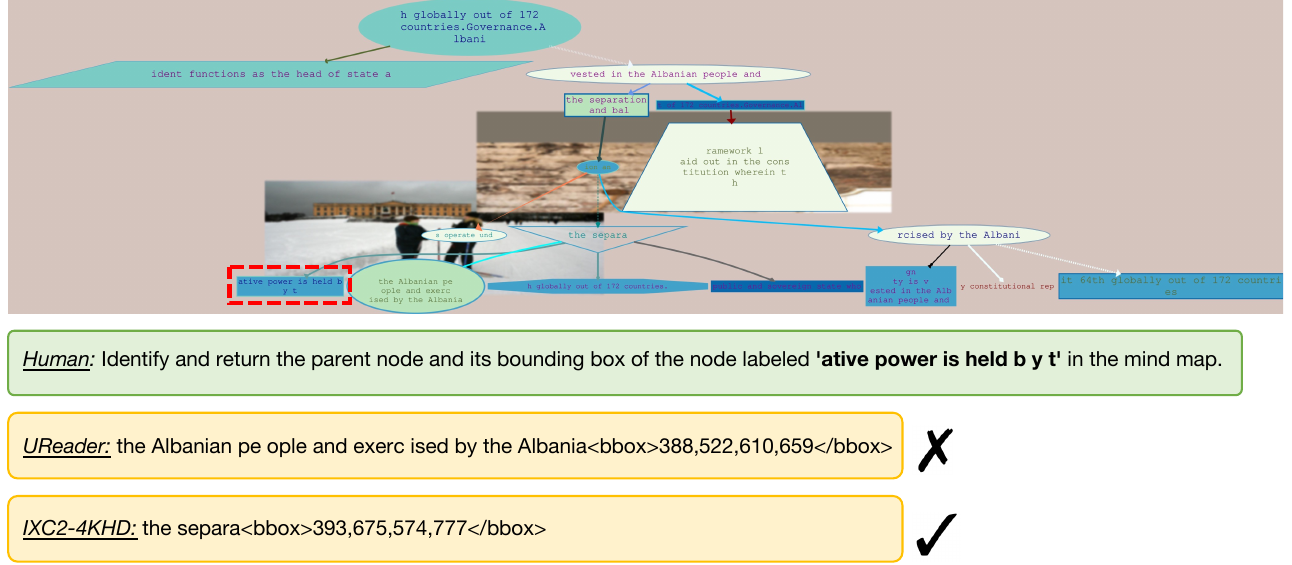}
        \caption{Position-related VQA}
        \label{fig:pred4}
    \end{subfigure}
    \caption{{A qualitative result of existing MLLMs tuned on the MindBench.}}
    \label{fig:sft_infer}
\vspace{-1mm}
\end{figure}

\subsection{Qualitative Results}
We first investigate the structured parsing capability of existing MLLMs through zero-shot inference, as depicted in Fig.~\ref{fig:zero-shot}. It is evident that GPT4V exhibits superior parsing ability. However, when confronted with closely positioned nodes, it tends to assign child nodes to incorrect parent nodes. This behavior can be attributed to the model's inclination to rely on layout information rather than inter-node interactions for determining node relationships. On the other hand, IXC2-4KHD demonstrates weaker zero-shot parsing ability. While the model comprehends the markdown format in the prompt, it can only generate flat prediction results with incomplete texts.

Next, we present the prediction results of UReader and IXC2-4KHD tuned on the MindBench, as depicted in Fig.~\ref{fig:sft_infer}. It is evident that IXC2-4KHD outperforms UReader across all four tasks, showcasing its strengths in comprehending node interactions, spatial perception, and structure parsing. In Fig.~\ref{fig:pred2}, IXC2-4KHD can successfully correlate spatial information with subgraph structure; however, it still faces challenges in parsing details, such as recognizing small text and accurately determining parent-child relationships.

\section{Conclusion}
\label{sec:conclusion}
In this paper, we introduce MindBench, the first comprehensive benchmark designed for structured document. MindBench stands out due to two primary features: 1) abundant structured document images with detailed annotations and evaluation metrics, providing a standardized research tool; 2) unified structure learning of five mind map understanding and parsing tasks that comprehensively assess the model's ability to text recognition, spatial awareness, relationship discernment, and structured parsing. We empirically investigate multiple visual document understanding baseline methods on the MindBench dataset. Experimental results demonstrate that there is significant room for improvement in current models' performance, particularly in handling high-resolution complex images and processing lengthy structured documents.

{\flushleft \bf Future work.} This paper primarily focuses on establishing a benchmark for structured document parsing of mind maps. Although the data sources include various styles such as tables, relationship diagrams, and posters, mind map data predominates. In the future, we aim to expand structured document parsing to encompass a wider range of graphical types, enabling the understanding of information in any graphical document.

\begin{ack}
We extend our heartfelt appreciation to XMind, Biggerplate, and Zhixi website for providing open-source mind map data, which played a crucial role in organizing this dataset.
\end{ack}

{\small
\bibliographystyle{plain}
\bibliography{ref}

\begin{thebibliography}{10}

\bibitem{alayrac2022flamingo}
Jean-Baptiste Alayrac, Jeff Donahue, Pauline Luc, Antoine Miech, Iain Barr, Yana Hasson, Karel Lenc, Arthur Mensch, Katherine Millican, Malcolm Reynolds, et~al.
\newblock Flamingo: a visual language model for few-shot learning.
\newblock {\em Advances in Neural Information Processing Systems}, 2022.

\bibitem{appalaraju2024docformerv2}
Srikar Appalaraju, Peng Tang, Qi~Dong, Nishant Sankaran, Yichu Zhou, and R~Manmatha.
\newblock Docformerv2: Local features for document understanding.
\newblock In {\em Proceedings of the AAAI Conference on Artificial Intelligence}, 2024.

\bibitem{bai2023qwen}
Jinze Bai, Shuai Bai, Shusheng Yang, Shijie Wang, Sinan Tan, Peng Wang, Junyang Lin, Chang Zhou, and Jingren Zhou.
\newblock Qwen-vl: A versatile vision-language model for understanding, localization, text reading, and beyond.
\newblock {\em arXiv preprint arXiv:2308.12966}, 2023.

\bibitem{brown2020language}
Tom Brown, Benjamin Mann, Nick Ryder, Melanie Subbiah, Jared~D Kaplan, Prafulla Dhariwal, Arvind Neelakantan, Pranav Shyam, Girish Sastry, Amanda Askell, et~al.
\newblock Language models are few-shot learners.
\newblock {\em Advances in Neural Information Processing Systems}, 2020.

\bibitem{chen2019tabfact}
Wenhu Chen, Hongmin Wang, Jianshu Chen, Yunkai Zhang, Hong Wang, Shiyang Li, Xiyou Zhou, and William~Yang Wang.
\newblock Tabfact: A large-scale dataset for table-based fact verification.
\newblock {\em arXiv preprint arXiv:1909.02164}, 2019.

\bibitem{chen2021websrc}
Xingyu Chen, Zihan Zhao, Lu~Chen, Danyang Zhang, Jiabao Ji, Ao~Luo, Yuxuan Xiong, and Kai Yu.
\newblock Websrc: A dataset for web-based structural reading comprehension.
\newblock {\em arXiv preprint arXiv:2101.09465}, 2021.

\bibitem{chen2024far}
Zhe Chen, Weiyun Wang, Hao Tian, Shenglong Ye, Zhangwei Gao, Erfei Cui, Wenwen Tong, Kongzhi Hu, Jiapeng Luo, Zheng Ma, et~al.
\newblock How far are we to gpt-4v? closing the gap to commercial multimodal models with open-source suites.
\newblock {\em arXiv preprint arXiv:2404.16821}, 2024.

\bibitem{chen2023internvl}
Zhe Chen, Jiannan Wu, Wenhai Wang, Weijie Su, Guo Chen, Sen Xing, Zhong Muyan, Qinglong Zhang, Xizhou Zhu, Lewei Lu, et~al.
\newblock Internvl: Scaling up vision foundation models and aligning for generic visual-linguistic tasks.
\newblock {\em arXiv preprint arXiv:2312.14238}, 2023.

\bibitem{chi2019complicated}
Zewen Chi, Heyan Huang, Heng-Da Xu, Houjin Yu, Wanxuan Yin, and Xian-Ling Mao.
\newblock Complicated table structure recognition.
\newblock {\em arXiv preprint arXiv:1908.04729}, 2019.

\bibitem{ch2017total}
Chee~Kheng Ch'ng and Chee~Seng Chan.
\newblock Total-text: A comprehensive dataset for scene text detection and recognition.
\newblock In {\em 2017 14th IAPR international conference on document analysis and recognition (ICDAR)}, 2017.

\bibitem{davis2022end}
Brian Davis, Bryan Morse, Brian Price, Chris Tensmeyer, Curtis Wigington, and Vlad Morariu.
\newblock End-to-end document recognition and understanding with dessurt.
\newblock In {\em European Conference on Computer Vision}, 2022.

\bibitem{dong2024internlm4k}
Xiaoyi Dong, Pan Zhang, Yuhang Zang, Yuhang Cao, Bin Wang, Linke Ouyang, Songyang Zhang, Haodong Duan, Wenwei Zhang, Yining Li, et~al.
\newblock Internlm-xcomposer2-4khd: A pioneering large vision-language model handling resolutions from 336 pixels to 4k hd.
\newblock {\em arXiv preprint arXiv:2404.06512}, 2024.

\bibitem{dosovitskiy2020image}
Alexey Dosovitskiy, Lucas Beyer, Alexander Kolesnikov, Dirk Weissenborn, Xiaohua Zhai, Thomas Unterthiner, Mostafa Dehghani, Matthias Minderer, Georg Heigold, Sylvain Gelly, et~al.
\newblock An image is worth 16x16 words: Transformers for image recognition at scale.
\newblock {\em arXiv preprint arXiv:2010.11929}, 2020.

\bibitem{feng2023docpedia}
Hao Feng, Qi~Liu, Hao Liu, Wengang Zhou, Houqiang Li, and Can Huang.
\newblock Docpedia: Unleashing the power of large multimodal model in the frequency domain for versatile document understanding.
\newblock {\em arXiv preprint arXiv:2311.11810}, 2023.

\bibitem{feng2023unidoc}
Hao Feng, Zijian Wang, Jingqun Tang, Jinghui Lu, Wengang Zhou, Houqiang Li, and Can Huang.
\newblock Unidoc: A universal large multimodal model for simultaneous text detection, recognition, spotting and understanding.
\newblock {\em arXiv preprint arXiv:2308.11592}, 2023.

\bibitem{hong2023cogagent}
Wenyi Hong, Weihan Wang, Qingsong Lv, Jiazheng Xu, Wenmeng Yu, Junhui Ji, Yan Wang, Zihan Wang, Yuxiao Dong, Ming Ding, et~al.
\newblock Cogagent: A visual language model for gui agents.
\newblock {\em arXiv preprint arXiv:2312.08914}, 2023.

\bibitem{hu2021question}
Anwen Hu, Shizhe Chen, and Qin Jin.
\newblock Question-controlled text-aware image captioning.
\newblock In {\em ACM International conference on Multimedia}, 2021.

\bibitem{hu2023mplug}
Anwen Hu, Yaya Shi, Haiyang Xu, Jiabo Ye, Qinghao Ye, Ming Yan, Chenliang Li, Qi~Qian, Ji~Zhang, and Fei Huang.
\newblock mplug-paperowl: Scientific diagram analysis with the multimodal large language model.
\newblock {\em arXiv preprint arXiv:2311.18248}, 2023.

\bibitem{hu2024mplug}
Anwen Hu, Haiyang Xu, Jiabo Ye, Ming Yan, Liang Zhang, Bo~Zhang, Chen Li, Ji~Zhang, Qin Jin, Fei Huang, et~al.
\newblock mplug-docowl 1.5: Unified structure learning for ocr-free document understanding.
\newblock {\em arXiv preprint arXiv:2403.12895}, 2024.

\bibitem{huang2022layoutlmv3}
Yupan Huang, Tengchao Lv, Lei Cui, Yutong Lu, and Furu Wei.
\newblock Layoutlmv3: Pre-training for document ai with unified text and image masking.
\newblock In {\em Proceedings of the 30th ACM International Conference on Multimedia}, 2022.

\bibitem{huang2019icdar2019}
Zheng Huang, Kai Chen, Jianhua He, Xiang Bai, Dimosthenis Karatzas, Shijian Lu, and CV~Jawahar.
\newblock Icdar2019 competition on scanned receipt ocr and information extraction.
\newblock In {\em 2019 International Conference on Document Analysis and Recognition (ICDAR)}, 2019.

\bibitem{jaume2019funsd}
Guillaume Jaume, Hazim~Kemal Ekenel, and Jean-Philippe Thiran.
\newblock Funsd: A dataset for form understanding in noisy scanned documents.
\newblock In {\em 2019 International Conference on Document Analysis and Recognition Workshops (ICDARW)}, 2019.

\bibitem{kafle2018dvqa}
Kushal Kafle, Brian Price, Scott Cohen, and Christopher Kanan.
\newblock Dvqa: Understanding data visualizations via question answering.
\newblock In {\em IEEE/CVF Conference on Computer Vision and Pattern Recognition}, 2018.

\bibitem{kantharaj2022chart}
Shankar Kantharaj, Rixie Tiffany~Ko Leong, Xiang Lin, Ahmed Masry, Megh Thakkar, Enamul Hoque, and Shafiq Joty.
\newblock Chart-to-text: A large-scale benchmark for chart summarization.
\newblock {\em arXiv preprint arXiv:2203.06486}, 2022.

\bibitem{kim2022ocr}
Geewook Kim, Teakgyu Hong, Moonbin Yim, JeongYeon Nam, Jinyoung Park, Jinyeong Yim, Wonseok Hwang, Sangdoo Yun, Dongyoon Han, and Seunghyun Park.
\newblock Ocr-free document understanding transformer.
\newblock In {\em European Conference on Computer Vision}, 2022.

\bibitem{lee2023pix2struct}
Kenton Lee, Mandar Joshi, Iulia~Raluca Turc, Hexiang Hu, Fangyu Liu, Julian~Martin Eisenschlos, Urvashi Khandelwal, Peter Shaw, Ming-Wei Chang, and Kristina Toutanova.
\newblock Pix2struct: Screenshot parsing as pretraining for visual language understanding.
\newblock In {\em International Conference on Machine Learning}, 2023.

\bibitem{li2023blip}
Junnan Li, Dongxu Li, Silvio Savarese, and Steven Hoi.
\newblock Blip-2: Bootstrapping language-image pre-training with frozen image encoders and large language models.
\newblock In {\em International Conference on Machine Learning}, 2023.

\bibitem{li2023monkey}
Zhang Li, Biao Yang, Qiang Liu, Zhiyin Ma, Shuo Zhang, Jingxu Yang, Yabo Sun, Yuliang Liu, and Xiang Bai.
\newblock Monkey: Image resolution and text label are important things for large multi-modal models.
\newblock {\em arXiv preprint arXiv:2311.06607}, 2023.

\bibitem{liu2023improved}
Haotian Liu, Chunyuan Li, Yuheng Li, and Yong~Jae Lee.
\newblock Improved baselines with visual instruction tuning.
\newblock {\em arXiv preprint arXiv:2310.03744}, 2023.

\bibitem{liu2024visual}
Haotian Liu, Chunyuan Li, Qingyang Wu, and Yong~Jae Lee.
\newblock Visual instruction tuning.
\newblock {\em Advances in Neural Information Processing Systems}, 2024.

\bibitem{liu2019curved}
Yuliang Liu, Lianwen Jin, Shuaitao Zhang, Canjie Luo, and Sheng Zhang.
\newblock Curved scene text detection via transverse and longitudinal sequence connection.
\newblock {\em Pattern Recognition}, 2019.

\bibitem{liu2024textmonkey}
Yuliang Liu, Biao Yang, Qiang Liu, Zhang Li, Zhiyin Ma, Shuo Zhang, and Xiang Bai.
\newblock Textmonkey: An ocr-free large multimodal model for understanding document.
\newblock {\em arXiv preprint arXiv:2403.04473}, 2024.

\bibitem{long2021parsing}
Rujiao Long, Wen Wang, Nan Xue, Feiyu Gao, Zhibo Yang, Yongpan Wang, and Gui-Song Xia.
\newblock Parsing table structures in the wild.
\newblock In {\em IEEE/CVF International Conference on Computer Vision}, 2021.

\bibitem{mao2024visually}
Zhiming Mao, Haoli Bai, Lu~Hou, Jiansheng Wei, Xin Jiang, Qun Liu, and Kam-Fai Wong.
\newblock Visually guided generative text-layout pre-training for document intelligence.
\newblock {\em arXiv preprint arXiv:2403.16516}, 2024.

\bibitem{masry2022chartqa}
Ahmed Masry, Do~Xuan Long, Jia~Qing Tan, Shafiq Joty, and Enamul Hoque.
\newblock Chartqa: A benchmark for question answering about charts with visual and logical reasoning.
\newblock {\em arXiv preprint arXiv:2203.10244}, 2022.

\bibitem{mathew2022infographicvqa}
Minesh Mathew, Viraj Bagal, Rub{\`e}n Tito, Dimosthenis Karatzas, Ernest Valveny, and CV~Jawahar.
\newblock Infographicvqa.
\newblock In {\em Proceedings of the IEEE/CVF Winter Conference on Applications of Computer Vision}, 2022.

\bibitem{mathew2021docvqa}
Minesh Mathew, Dimosthenis Karatzas, and CV~Jawahar.
\newblock Docvqa: A dataset for vqa on document images.
\newblock In {\em Proceedings of the IEEE/CVF winter conference on applications of computer vision}, 2021.

\bibitem{methani2020plotqa}
Nitesh Methani, Pritha Ganguly, Mitesh~M Khapra, and Pratyush Kumar.
\newblock Plotqa: Reasoning over scientific plots.
\newblock In {\em Proceedings of the IEEE/CVF Winter Conference on Applications of Computer Vision}, 2020.

\bibitem{openai2023gpt}
R~OpenAI.
\newblock Gpt-4 technical report. arxiv 2303.08774.
\newblock {\em View in Article}, 2023.

\bibitem{park2019cord}
Seunghyun Park, Seung Shin, Bado Lee, Junyeop Lee, Jaeheung Surh, Minjoon Seo, and Hwalsuk Lee.
\newblock Cord: a consolidated receipt dataset for post-ocr parsing.
\newblock In {\em Workshop on Document Intelligence at NeurIPS 2019}, 2019.

\bibitem{pasupat2015compositional}
Panupong Pasupat and Percy Liang.
\newblock Compositional semantic parsing on semi-structured tables.
\newblock {\em arXiv preprint arXiv:1508.00305}, 2015.

\bibitem{radford2021learning}
Alec Radford, Jong~Wook Kim, Chris Hallacy, Aditya Ramesh, Gabriel Goh, Sandhini Agarwal, Girish Sastry, Amanda Askell, Pamela Mishkin, Jack Clark, et~al.
\newblock Learning transferable visual models from natural language supervision.
\newblock In {\em International Conference on Machine Learning}, 2021.

\bibitem{riba2019table}
Pau Riba, Anjan Dutta, Lutz Goldmann, Alicia Forn{\'e}s, Oriol Ramos, and Josep Llad{\'o}s.
\newblock Table detection in invoice documents by graph neural networks.
\newblock In {\em 2019 International Conference on Document Analysis and Recognition (ICDAR)}, 2019.

\bibitem{sidorov2020textcaps}
Oleksii Sidorov, Ronghang Hu, Marcus Rohrbach, and Amanpreet Singh.
\newblock Textcaps: a dataset for image captioning with reading comprehension.
\newblock In {\em European Conference on Computer Vision}, 2020.

\bibitem{singh2019towards}
Amanpreet Singh, Vivek Natarajan, Meet Shah, Yu~Jiang, Xinlei Chen, Dhruv Batra, Devi Parikh, and Marcus Rohrbach.
\newblock Towards vqa models that can read.
\newblock In {\em IEEE/CVF Conference on Computer Vision and Pattern Recognition}, 2019.

\bibitem{stanislawek2021kleister}
Tomasz Stanis{\l}awek, Filip Grali{\'n}ski, Anna Wr{\'o}blewska, Dawid Lipi{\'n}ski, Agnieszka Kaliska, Paulina Rosalska, Bartosz Topolski, and Przemys{\l}aw Biecek.
\newblock Kleister: key information extraction datasets involving long documents with complex layouts.
\newblock In {\em International Conference on Document Analysis and Recognition}, 2021.

\bibitem{svetlichnaya2020deepform}
S~Svetlichnaya.
\newblock Deepform: Understand structured documents at scale, 2020.

\bibitem{tanaka2021visualmrc}
Ryota Tanaka, Kyosuke Nishida, and Sen Yoshida.
\newblock Visualmrc: Machine reading comprehension on document images.
\newblock In {\em Proceedings of the AAAI Conference on Artificial Intelligence}, 2021.

\bibitem{tang2023vistext}
Benny~J Tang, Angie Boggust, and Arvind Satyanarayan.
\newblock Vistext: A benchmark for semantically rich chart captioning.
\newblock {\em arXiv preprint arXiv:2307.05356}, 2023.

\bibitem{tang2023unifying}
Zineng Tang, Ziyi Yang, Guoxin Wang, Yuwei Fang, Yang Liu, Chenguang Zhu, Michael Zeng, Cha Zhang, and Mohit Bansal.
\newblock Unifying vision, text, and layout for universal document processing.
\newblock In {\em IEEE/CVF Conference on Computer Vision and Pattern Recognition}, 2023.

\bibitem{touvron2023llama}
Hugo Touvron, Thibaut Lavril, Gautier Izacard, Xavier Martinet, Marie-Anne Lachaux, Timoth{\'e}e Lacroix, Baptiste Rozi{\`e}re, Naman Goyal, Eric Hambro, Faisal Azhar, et~al.
\newblock Llama: Open and efficient foundation language models.
\newblock {\em arXiv preprint arXiv:2302.13971}, 2023.

\bibitem{vicuna}
Vicuna.
\newblock Vicuna: An open chatbot impressing gpt-4.
\newblock \url{https://github.com/lm-sys/FastChat}, 2023.

\bibitem{wan2024omniparser}
Jianqiang Wan, Sibo Song, Wenwen Yu, Yuliang Liu, Wenqing Cheng, Fei Huang, Xiang Bai, Cong Yao, and Zhibo Yang.
\newblock Omniparser: A unified framework for text spotting, key information extraction and table recognition.
\newblock {\em arXiv preprint arXiv:2403.19128}, 2024.

\bibitem{wang2023docllm}
Dongsheng Wang, Natraj Raman, Mathieu Sibue, Zhiqiang Ma, Petr Babkin, Simerjot Kaur, Yulong Pei, Armineh Nourbakhsh, and Xiaomo Liu.
\newblock Docllm: A layout-aware generative language model for multimodal document understanding.
\newblock {\em arXiv preprint arXiv:2401.00908}, 2023.

\bibitem{xu2020layoutlmv2}
Yang Xu, Yiheng Xu, Tengchao Lv, Lei Cui, Furu Wei, Guoxin Wang, Yijuan Lu, Dinei Florencio, Cha Zhang, Wanxiang Che, et~al.
\newblock Layoutlmv2: Multi-modal pre-training for visually-rich document understanding.
\newblock {\em arXiv preprint arXiv:2012.14740}, 2020.

\bibitem{xu2020layoutlm}
Yiheng Xu, Minghao Li, Lei Cui, Shaohan Huang, Furu Wei, and Ming Zhou.
\newblock Layoutlm: Pre-training of text and layout for document image understanding.
\newblock In {\em Proceedings of the 26th ACM SIGKDD international conference on knowledge discovery \& data mining}, 2020.

\bibitem{ye2023mplugdoc}
Jiabo Ye, Anwen Hu, Haiyang Xu, Qinghao Ye, Ming Yan, Yuhao Dan, Chenlin Zhao, Guohai Xu, Chenliang Li, Junfeng Tian, et~al.
\newblock mplug-docowl: Modularized multimodal large language model for document understanding.
\newblock {\em arXiv preprint arXiv:2307.02499}, 2023.

\bibitem{ye2023ureader}
Jiabo Ye, Anwen Hu, Haiyang Xu, Qinghao Ye, Ming Yan, Guohai Xu, Chenliang Li, Junfeng Tian, Qi~Qian, Ji~Zhang, et~al.
\newblock Ureader: Universal ocr-free visually-situated language understanding with multimodal large language model.
\newblock {\em arXiv preprint arXiv:2310.05126}, 2023.

\bibitem{ye2023mplug}
Qinghao Ye, Haiyang Xu, Guohai Xu, Jiabo Ye, Ming Yan, Yiyang Zhou, Junyang Wang, Anwen Hu, Pengcheng Shi, Yaya Shi, et~al.
\newblock mplug-owl: Modularization empowers large language models with multimodality.
\newblock {\em arXiv preprint arXiv:2304.14178}, 2023.

\bibitem{ye2023mplug2}
Qinghao Ye, Haiyang Xu, Jiabo Ye, Ming Yan, Haowei Liu, Qi~Qian, Ji~Zhang, Fei Huang, and Jingren Zhou.
\newblock mplug-owl2: Revolutionizing multi-modal large language model with modality collaboration.
\newblock {\em arXiv preprint arXiv:2311.04257}, 2023.

\bibitem{yin2023survey}
Shukang Yin, Chaoyou Fu, Sirui Zhao, Ke~Li, Xing Sun, Tong Xu, and Enhong Chen.
\newblock A survey on multimodal large language models.
\newblock {\em arXiv preprint arXiv:2306.13549}, 2023.

\bibitem{yuan2022syntax}
Ye~Yuan, Xiao Liu, Wondimu Dikubab, Hui Liu, Zhilong Ji, Zhongqin Wu, and Xiang Bai.
\newblock Syntax-aware network for handwritten mathematical expression recognition.
\newblock In {\em IEEE/CVF Conference on Computer Vision and Pattern Recognition}, 2022.

\bibitem{Zhang_Shasha_1989}
Kaizhong Zhang and Dennis Shasha.
\newblock Simple fast algorithms for the editing distance between trees and related problems.
\newblock {\em SIAM Journal on Computing}, 1989.

\bibitem{zhang2023llavar}
Yanzhe Zhang, Ruiyi Zhang, Jiuxiang Gu, Yufan Zhou, Nedim Lipka, Diyi Yang, and Tong Sun.
\newblock Llavar: Enhanced visual instruction tuning for text-rich image understanding.
\newblock {\em arXiv preprint arXiv:2306.17107}, 2023.

\bibitem{zhao2023survey}
Wayne~Xin Zhao, Kun Zhou, Junyi Li, Tianyi Tang, Xiaolei Wang, Yupeng Hou, Yingqian Min, Beichen Zhang, Junjie Zhang, Zican Dong, et~al.
\newblock A survey of large language models.
\newblock {\em arXiv preprint arXiv:2303.18223}, 2023.

\bibitem{zhong2020image}
Xu~Zhong, Elaheh ShafieiBavani, and Antonio Jimeno~Yepes.
\newblock Image-based table recognition: data, model, and evaluation.
\newblock In {\em European Conference on Computer Vision}, 2020.

\bibitem{zhu2023minigpt}
Deyao Zhu, Jun Chen, Xiaoqian Shen, Xiang Li, and Mohamed Elhoseiny.
\newblock Minigpt-4: Enhancing vision-language understanding with advanced large language models.
\newblock {\em arXiv preprint arXiv:2304.10592}, 2023.

\end{thebibliography}
}

\end{document}